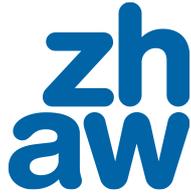

**Zurich University of Applied Sciences**

School of Engineering

Institut für Informatik (InIT)

Bachelor's Thesis

# Focusing on Students, not Machines: Grounded Question Generation and Automated Answer Grading

*Authors:*
Philip Breuer[1]
Gérôme Meyer[2]

*Supervisor:*
Dr. Jonathan Fürst

Submitted on
June 7, 2024

Study program:
[1]Informatik, B.Sc.
[2]Data Science, B.Sc.

# Imprint

# DECLARATION OF ORIGINALITY

# Bachelor thesis at the School of Engineering

## Declaration of originality

I hereby declare that I have written this thesis independently or together with the listed group members.

I have only used the sources and aids (including websites and generative AI tools) specified in the text or appendix. I am responsible for the quality of the text and the selection of all content and have ensured that information and arguments are substantiated or supported by appropriate scientific sources. Generative AI tools have been summarized by name and purpose.

Any misconduct will be dealt with according to paragraphs 39 and 40 of the General Academic Regulations for Bachelor's and Master's Degree courses at the Zurich University of Applied Sciences (Rahmenprüfungsordnung ZHAW (RPO)) and subject to the provisions for disciplinary action stipulated in the University regulations.

**Ort, Datum:**                              **Name Studierende:**

Winterthur, 07. June 2024                    Philip Breuer

                                             Gérôme Meyer

# Abstract

Digital technologies are increasingly used in education to reduce the workload of teachers and students. However, creating open-ended study or examination questions and grading their answers is still a tedious task. This thesis presents the foundation for a system that generates questions grounded in class materials and automatically grades student answers. It introduces a sophisticated method for chunking documents with a visual layout, specifically targeting PDF documents. This method enhances the accuracy of downstream tasks, including Retrieval Augmented Generation (RAG). Our thesis demonstrates that high-quality questions and reference answers can be generated from study material. Further, it introduces a new benchmark for automated grading of short answers to facilitate comparison of automated grading systems. An evaluation of various grading systems is conducted and indicates that Large Language Models (LLMs) can generalise to the task of automated grading of short answers from their pre-training tasks. As with other tasks, increasing the parameter size of the LLMs leads to greater performance. Currently, available systems still need human oversight, especially in examination scenarios.

# Zusammenfassung

Digitale Technologien werden immer mehr im Bildungsbereich angewendet, um die Arbeitslast von Lehrkräften und Lernenden zu reduzieren. Das Erstellen und Benoten von Übungs- und Prüfungsfragen ist bis anhin eine aufwändige Aufgabe. Die vorliegende Arbeit präsentiert die Grundlage für ein System, welches Fragen basierend auf Unterrichtsmaterialien generiert und automatisch Antworten zu diesen Fragen benoten kann. Sie führt eine neue Methode ein, um Unterrichtsmaterialien, spezifisch im PDF Format, anhand eines visuellen Modells in zusammenhänge Blöcke zu klassifizieren. Die vorgestellte Methode führt zu deutlich besseren Fragen, welche anhand vom Unterrichtsmaterial generiert werden. Des Weiteren wird ein neuer Datensatz zur Evaluierung von automatisierten Benotungssystemen zur Verfügung gestellt. Auf diesem Datensatz werden drei Arten von Systemen überprüft und daraus geht hervor, dass grosse Sprachmodelle auf diese Aufgabe hinaus generalisieren können. Weiter zeigt sich, das Sprachmodelle mit höherer Parameteranzahl zu besseren Resultaten tendieren, aber bisher noch nicht vollständig ohne menschliche Beaufsichtigung im Bildungsbereich angewendet werden sollten.

# Preface

This Bachelor's thesis was conducted at the Institute for Information Technology at the Zurich University for Applied Sciences ZHAW.

We would like to thank Jonathan Fürst for his feedback and guidance during the writing of this thesis.

This thesis was funded by OpenAI's Researcher Access Program.

# Contents

# 1. Introduction

Digital technologies have a wide range of applications in the field of education. From equipping classrooms with computers to freely accessible online learning platforms that are used completely independently of traditional educational institutions. There is an ever-growing number of teachers and students who use digital tools [1]. One of the main motivations for this digitalisation of education is to reduce the administrative workload for teaching professionals and students alike.

P. C. Brown, H. L. Roediger III, and M. A. McDaniel [2] summarise the scientific consensus on learning and conclude that passively reading information without active engagement with the content has very few long-term benefits. They argue that only reading about information leads to poor retention and little understanding of concepts and ideas. Questions are a good way of getting students to engage with a topic but are not used as often as they could be because creating relevant questions on a topic is a time-intensive task.

Before the invention of **LLMs**, automated generation of various types of questions was not feasible. Since then, commercial tools have started providing limited question-generation functionality.

However, simply answering questions does not provide much benefit on its own. Students require some form of feedback to be able to spot and correct mistakes and misunderstandings. Often, this feedback, especially for examinations, comes in the form of a grade. Grading also requires time and is usually very repetitive. Hence, when questions are used during learning, they tend to be closed-ended with predefined answers that students have to pick from. This is despite research showing that these types of questions do not adequately test students' understanding and often promote short-term memorisation. One of the reasons for the popularity of closed-ended questions is that they are easier to grade [2].

In response to this, automated grading of student answers has become an active field of research [3]–[9]. Existing educational tools often include heuristic algorithms such as exact text matching or fuzzy matching to automatically grade answers. However, these approaches struggle to grade correct answers containing grammatical mistakes or different formulations of the same meaning. Because of these limitations, such systems make mistakes that frustrate users. This leads to questions being adjusted so that they conform to the algorithms' grading capabilities. Essentially, this leads to questions being turned back into closed-ended ones [2].

This thesis aims to reduce the friction for students and teachers to use open-ended questions during learning. Recently, **LLMs** were shown to have extensive knowledge about various topics and can generate new questions about them [10]. However, study sessions and examinations do not happen in isolation, so questions need to be grounded in topics covered during class. To achieve this, an **RAG** approach is applied to extract information from study materials.

## 1.1. Contribution

We introduce the combined ASAG2024[1] benchmark to facilitate the comparison of automated grading systems. Further, we present initial evaluations of existing automated grading solutions on the benchmark. We show that specialised grading systems are still limited in their ability to generalise to new questions and may need to be fine-tuned for specific use cases. **LLMs** are shown to be able to generalise to the grading task with decent performance without being specifically trained or fine-tuned to any specific grading data. Aligned with related research, as the size of an **LLM** is increased, its ability to generalise to the grading task improves [11].

Additionally, we introduce a sophisticated method for chunking documents with a visual layout, specifically targeting PDF documents. This method enhances the accuracy of downstream tasks, including **RAG**. Our thesis demonstrates that combining study materials, learning objectives, and instructions based on Bloom's taxonomy yields high-quality, relevant questions and reference answers.

---

[1]Available online: **https://huggingface.co/datasets/Meyerger/ASAG2024**

# 2. Background

In this section, background information is provided on the core concepts relevant to this thesis. First, **LLMs**, which are a central part of this thesis, are introduced and then follows a description of text embedding models, which convert text into numerical representations. Cosine similarity is a metric used to assess the similarity between these embeddings. Prompt Engineering describes multiple techniques for guiding **LLMs** towards desired outputs through specific instructions. Finally, the background section concludes with an explanation of Portable Document Format (PDF) and Bloom's Revised Taxonomy, both of which are relevant to the question generation described in this thesis.

## 2.1. Large Language Models

The term **Large Language Model (LLM)** seems to have been popularised by the work done at OpenAI. They demonstrated that language models can learn a variety of NLP tasks without the need for labelled data when scaled to large enough sizes. The two main components necessary for this advancement were the Transformer Neural architecture and a large corpus of sufficient quality [12]. These models are often trained in two separate stages. First, the model is trained on a large corpus of text with a language modelling objective and then adapted to a task on a smaller and more specific dataset [13]. The latter process is often called fine-tuning, resulting in a *fine-tuned* model, whereas the model trained in the first step is called a *foundational* model.

This thesis employs **LLMs** for the question generation step, described in **Section 4**, and evaluates their ability to grade student answers in **Section 5**.

## 2.2. Embedding Models

Text embedding models transform text or sentences into a higher dimensional vector space. These vectors capture the meaning of the embedded texts so that the distances and directions represent relationships between different sentences and words. This is achieved through training on large datasets where the model learns to position similar items closer together in the vector space according to how often they appear close to one another [14].

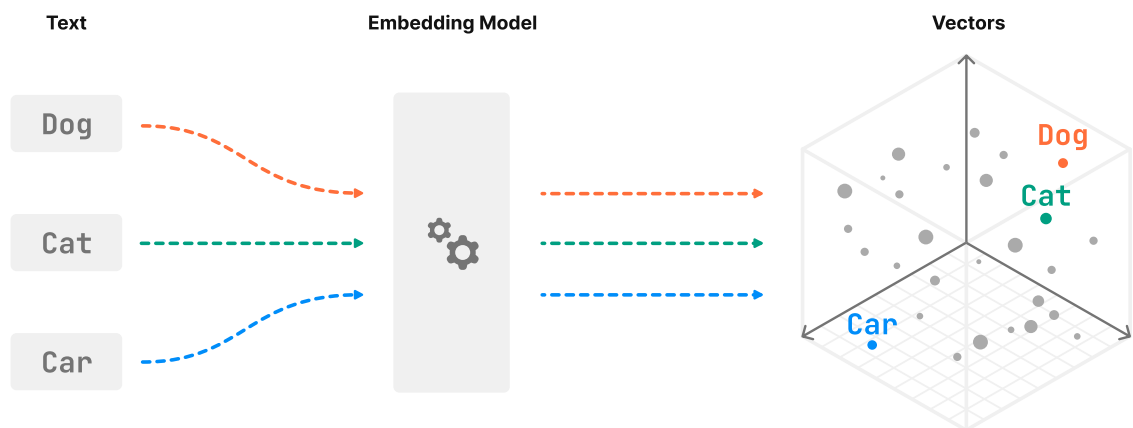

**Figure 1:** Illustration of text being translated into vectors via an embedding model.

In **Figure 1**, the words *Dog*, *Cat*, and *Car* are all embedded using such a model. The vector space on the right shows how the words *Dog* and *Cat* are grouped closer together than the word *Car*, even though the words *Cat* and *Car* are more similar in spelling. Important to note here is that the vector space of the embedding models has a lot more dimensions, usually between 768 and 4096 dimensions, rather than the 3D representation for this example. This semantic similarity between two embedding vectors can be calculated using the cosine similarity discussed in **Section 2.3**.

## 2.3. Cosine Similarity

$$\text{cosine similarity} := \cos(\theta) = \frac{A \cdot B}{\|A\|\|B\|}$$

Measures the cosine of the angle between two non-zero vectors $A$ and $B$. Unlike the more widely known Euclidian distance function, cosine similarity only considers the angle between the two vectors, not their magnitude.

Cosine similarity is used to compute semantic similarity. Sentences with very similar meanings should have high semantic similarity, while those with different meanings should have low similarity [14].

Mathematically speaking, the output range of this function falls between $-1$ and $1$ compared to $0$ and $\infty$ for a distance function. However, in practice, the similarities of two embedding vectors range from 0 to 1.

In other words, when two embedding vectors have a cosine similarity close to 0, the two underlying texts passed to the embedding model have different semantic meanings. And vice versa, if the cosine similarity is closer to 1, they have a similar semantic meaning.

Notably, for cosine similarity to be used as a measure of semantic similarity, embedding models need to be trained with regard to the cosine similarity [14], [15].

## 2.4. Prompt Engineering

Prompt engineering involves finding questions or instructions that guide language models to produce desired outputs through a trial-and-error process. There are many techniques, each with its strengths and weaknesses, but for this thesis, the most relevant techniques are *Zero- and Few-Shot Prompting*, *Chain-of-Thought* and *Retrieval Augmented Generation* [16].

### 2.4.1. Zero- vs Few-Shot Prompting

Zero-shot prompting is the basic technique of instructing a language model to execute novel tasks without extensive fine-tuning, and instead relying on a prompt [17]. Giving concrete examples of the novel task, is then referred to as Few-shot prompting, and can improve the performance.

### 2.4.2. Chain-of-thought (CoT)

**CoT** prompting is a technique designed to improve the reasoning capabilities of language models by structuring the output into a sequence of steps [18]. This approach helps the model to "think" through the problem methodically.

### 2.4.3. Retrieval Augmented Generation

**Retrieval Augmented Generation (RAG)** is a technique where information is provided alongside an instruction as the input (or prompt) to the **LLM**. While **LLMs** demonstrate knowledge and understanding, their static training data and their tendency to hallucinate hinder the accuracy required for many tasks. The models could be retrained to incorporate the required information; however, this process is expensive and slow. That's where **RAG** can provide a cost-effective way to artificially introduce knowledge into the model by including relevant text sections into the prompt while simultaneously improving factual accuracy [10], [19].

## 2.5. PDF

The **Portable Document Format (PDF)** is a widely used data format invented by Adobe Systems, intended for viewing and exchanging documents in a platform-independent way [20]. A notable characteristic of **PDFs** is that text is sorted by coordinates rather than in the natural reading flow from left to right, making it difficult to reconstruct the natural sequence of the text for extraction purposes. Moreover, **PDF** text can be encoded in various ways, often using embedded or custom fonts, complicating the decoding process and potentially leading to the loss of text because of artefacts or Unicode issues.

**PDFs** also commonly feature complex layouts, including multi-column text, tables, and embedded images, all of which require sophisticated parsing algorithms to interpret correctly. The various text encodings and the potential lack of Unicode support in older **PDFs** further lead to character mapping problems, complicating text extraction even more. Additionally, there is no guarantee that a **PDF** file contains any text at all; for example, a scanned document without an **Optical Character Recognition (OCR)** step may only consist of images, even though it was saved as a **PDF** file.

## 2.6. Bloom's Revised Taxonomy

Bloom's Revised Taxonomy is a framework for categorising educational goals, objectives, and standards [21]. It includes six levels of cognitive processes arranged from lower to higher-order thinking skills, as seen in **Table 1**.

This framework assists educators in progressively targeting and developing students' cognitive abilities, ensuring a comprehensive approach to learning and critical thinking.

| | | |
|---|---|---|
| 1 | **Remember** | Retrieve relevant knowledge from long-term memory |
| 2 | **Understand** | Construct meaning from instructional messages, including oral, written, and graphic communication |
| 3 | **Apply** | Carry out or use a procedure in a given situation |
| 4 | **Analyse** | Break material into its constituent parts and determine how the parts relate to one another and to an overall structure or purpose |
| 5 | **Evaluate** | Make judgements based on criteria and standards |
| 6 | **Create** | Put elements together to form a coherent or functional whole; reorganise elements into a new pattern or structure |

**Table 1:** Bloom's Revised Taxonomy [21]

# 3. System Design

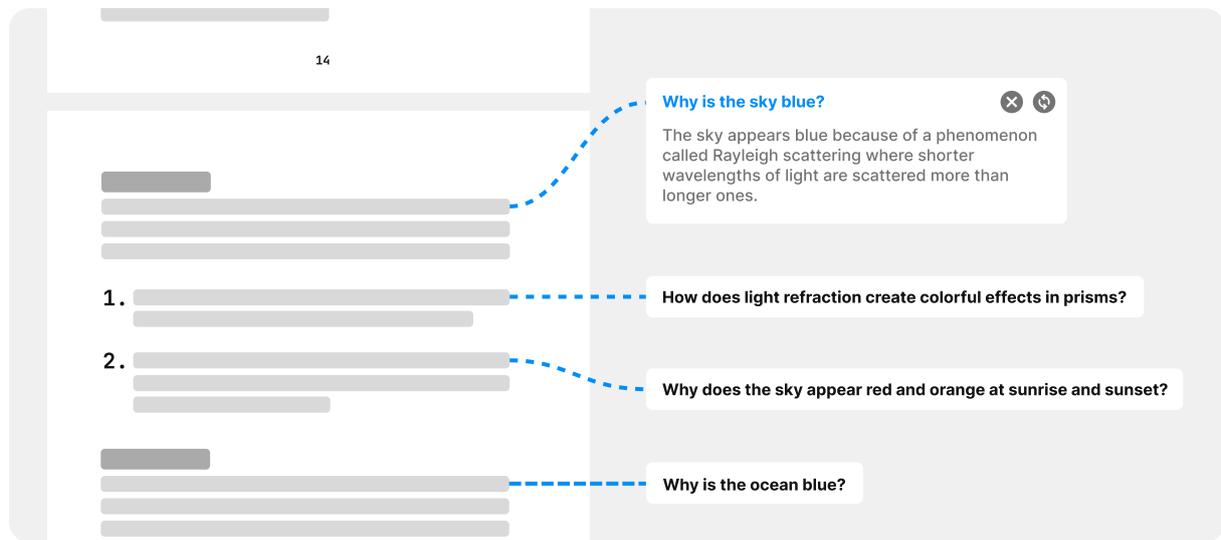

**Figure 2:** A mockup of the question generation user interface

This section provides an overview of the learning platform that influenced the design decisions for this thesis.

Existing learning platforms feature a straightforward, albeit time-consuming, approach. Students or Teachers can manually create flashcards with terms and definitions. After creating a list of cards, they can be answered and self-assessed. Depending on the assessment, an algorithm like **Spaced Repetition** influences how soon cards reappear in subsequent study sessions [2].

However, the static nature of flashcards limits the depth of the questions and emphasises memorisation over deeper understanding. Additionally, students are usually required to judge their answers themselves, which can be subjective and error-prone. While this approach works fine for expanding vocabulary in a foreign language or other simple memorisation tasks, assessments in schools and universities require knowledge that goes deeper than memorisation alone.

For that reason, the proposed platform aims to minimise the effort required to manually create study materials and the subjectivity of self-assessment while fostering a learning environment beyond memorisation.

This thesis focuses on the two key components of this system: question-and-answer generation and automated grading. The following sections will describe these two parts in detail. The final version of the learning platform will require additional components, such as a user interface, that were not the main focus and thus were excluded from the thesis.

## 3.1. Automated Question Generation

The proposed platform will allow students to upload documents containing course material, like slides, scripts, or exercise sheets. The platform will explicitly support **PDF** files, addressing the challenges that come with extracting information from this file format because it is commonly used to share these kinds of documents. The provided documents serve as the ground truth information from which the system derives questions and reference answers. This ensures that the generated questions are closely related to what students learn in their school or university. The questions should ideally only cover relevant content. Relevancy is difficult to assess just based on the documents alone. Instead of making assumptions, a list of topics and learning objectives is required for each topic alongside the documents. Ideally, these learning objectives outline the relevant sections and provide instructions about what knowledge or skills a student should acquire. It is not uncommon for schools and universities to share such learning objectives during lectures.

To make the questions more effective, Bloom's taxonomy is incorporated into the **Automatic Question Generation (AQG)** process as it is an essential framework for aligning with educational objectives. Specifically, the *Remember*, *Understand*, and *Apply* levels are particularly relevant because they involve questions that expect short, gradable answers. The higher-order levels of *Analyse*, *Evaluate*, and *Create*, while valuable, typically require more elaborate responses beyond the scope of the platform's capabilities. Thus, this thesis focuses on generating questions at the first three levels of Bloom's Taxonomy to ensure compatibility.

Each question will include information about its taxonomy level. This information will be utilised alongside **Spaced Repetition** to gradually increase the depth and difficulty of the questions as lower-level questions are mastered. This method is aimed at enhancing the learning experience and the effectiveness of study sessions.

An **LLM** is used to generate the questions and answers. Based on human evaluations performed by Elkins et al., **LLMs** are already capable of generating questions that are deemed useful for use in educational settings [22]. Nevertheless, these models require human oversight. Consequently, as can be seen in **Figure 2**, the user interface is designed to allow users to maintain control and verify the generated content.

How the questions and reference answers are generated is discussed in detail in **Section 4**.

## 3.2. Automated Grading

When it comes to the first three levels of Bloom's taxonomy, there are two ways of assessing a student's knowledge. Closed-ended questions, such as multiple-choice, provide a predefined set of answers from which the student must pick. Open-ended questions, on the other hand, require a full-text answer by the student. In general, open-ended questions require more effort by students and allow for a more in-depth evaluation of their knowledge. In a self-study environment, the additional effort required to write out answers has been shown to lead to better memorisation and understanding of the topic. However, the evaluation or self-assessment of answers to open-ended questions is naturally more difficult, time-consuming, and influenced by biases such as confirmation bias [2]. **Automated Short Answer Grading (ASAG)** has, therefore, the potential to free up teachers' time to focus on tutoring learners or adequately preparing classroom activities. Also, in self-study environments or online courses, where grading by a teacher is not feasible, such systems can provide valuable feedback and progress indicators to learners. Beyond the cost and time savings, reliable automated grading systems could open the door to new learning approaches, such as retrying an assignment as often as necessary and receiving immediate feedback. Additionally, from the student's perspective, being graded by a system can allow them to focus on their work without worrying about how their answer may be perceived by a teacher [2], [3], [5].

Various methods have been proven to work and are built into various existing systems to grade closed-ended questions automatically. Therefore, the focus is on assessing different methods for grading answer to open-ended questions from here on called short-answer grading.

Because of the potential benefits mentioned above, **Automated Short Answer Grading (ASAG)** has become a well-established field of research over the past few decades. The recent survey by E. del Gobbo et al. (2023) provided a comprehensive mapping of the literature regarding **ASAG** between 2016 and 2021. It concluded that **ASAG** systems were not ready for widespread general deployment yet and that the field lacked a comprehensive comparison of works. It notes that researchers often start from scratch when gathering relevant datasets and that these datasets are not uniform, making comparison between systems even more difficult.

To address these issues, **Section 5** introduces a new benchmark dataset for the ASAG task consisting of several commonly used datasets. Since new language processing techniques, such as Transformer Neural Networks and the rise of **LLMs**, have allowed for large strides in many similar fields between 2021 and 2024, **Section 5.2** investigates the performance of three types of grading approaches on the benchmark: "Semantic Similarity", "Specialised Grading Systems" and "Large Language Models" to assess whether any approaches have reached maturity and to which degree of autonomy they could be employed [3].

# 4. Question Generation

This section describes the steps for generating questions and their reference answers that are effective, grounded, accurate and relevant.

To ground questions in the study material, documents need to be fed to **LLMs**. However, since **LLMs** are constrained by specific context window sizes, documents cannot be passed in their entirety. Models with support for larger context windows exist, but research indicates that these models fail to effectively utilise and consider the entire context window, increasing the risk of hallucinations [23]. The status quo is to divide the documents into smaller, digestible chunks based on a heuristic algorithm.

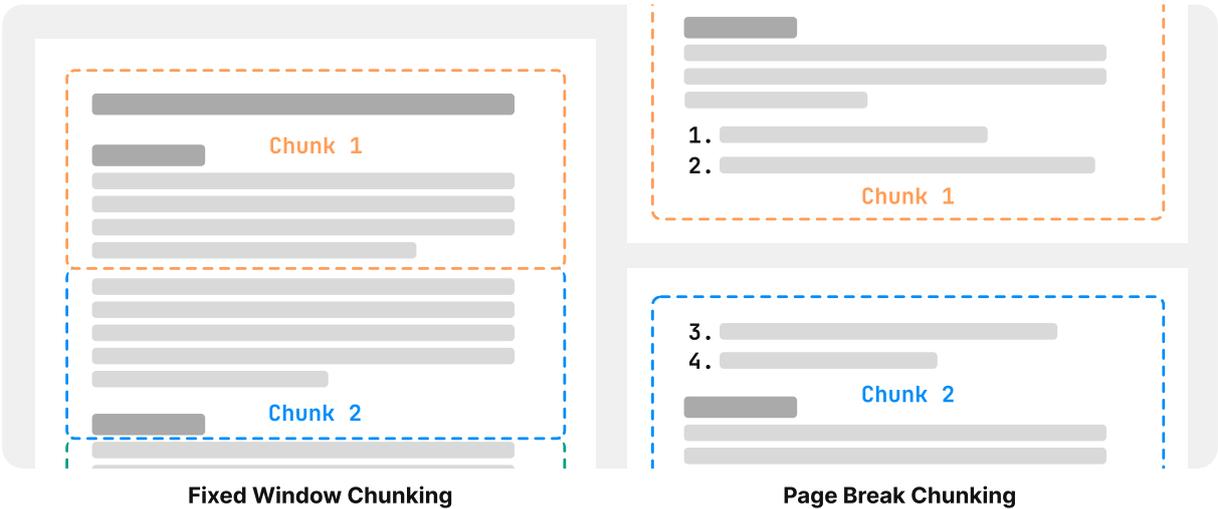

**Figure 3:** Illustration of context loss in fixed window and page break chunking

A naive but straightforward approach is to split a document based on a fixed-size window. This could be multiple blocks or entire pages. As seen on the left in **Figure 3**, using a small fixed-size window can lead to "context loss" by breaking apart semantically tightly coupled text sections that should stay together to preserve meaning.

Using page breaks as the window size does not eliminate the problem, as content regularly spills onto the next page. **Figure 3** contains an example on the right, where the first two list items are separated from the rest. From this point forward, these two issues will be referred to as "context loss" and "overflow".

There are ways to decrease the likelihood of losing context while using fixed window sizes, such as further increasing the window sizes, which defeats the purpose of chunking. Another option is to create overlapping chunks, which is not guaranteed to fix the issue and would generate duplicate questions in the later steps.

## 4.1. Methods

A more sophisticated and elegant approach is introduced in **Section 4.1.1** to ensure minimal context loss while still creating small enough chunks for the **LLM**. The resulting document chunks are then assigned to topics based on the learning objectives in **Section 4.1.2**. In the end, each chunk and the identified learning objectives are used to generate question-and-answer pairs in **Section 4.1.3**.

### 4.1.1. DLA Chunking

While text is the primary focus of most tasks extracting information from **PDF** documents, the format also contains visual information. The document's layout can reveal many insights into the semantic cohesion of text sections.

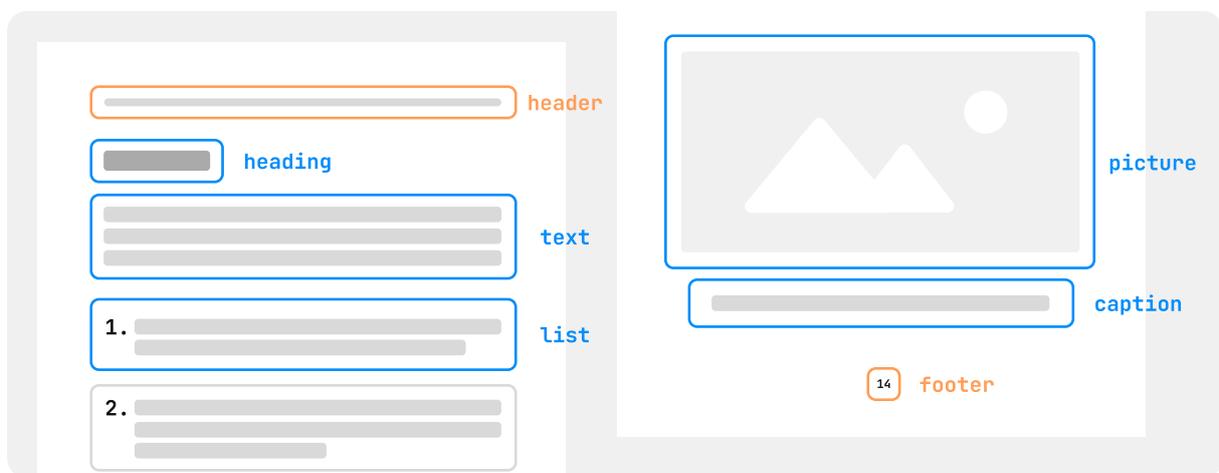

**Figure 4:** PDF blocks in grey overlaid with a few examples of layout structures, including their types and bounding boxes.

This requires a model that performs **Document Layout Analysis (DLA)** to extract the bounding boxes of individual layout structures like headings or paragraphs. Some examples can be seen in **Figure 4**. To distinguish these predicted layout structures from the **PDF** blocks, the term "instances" will be used instead. Multiple pre-trained models are available on the internet for this task. This thesis employs the popular YOLOv8[2] [24], [25] model fine-tuned on the DocLayNet dataset [26], [27].

Each predicted instance is associated with the existing blocks to determine the types for each block. The **DLA** model may create duplicate bounding boxes that fully or partially contain other instances. Using the intersection as an indicator for overlap would bias the associations towards the larger instance bounding boxes. Hence, the **Intersection over Union (IoU)** is used, as it effectively indicates overlap while accounting for size differences. Each block is compared to all instances on a page and is associated with the highest **IoU** value on the page.

Thanks to the identified layout structure types, the proposed chunking strategy can make assumptions on what blocks should stay together. For example, consecutive list items or tables are likely part of the same context. Ignoring `header` and `footer` types, these consecutive blocks can also be traced across page breaks, further preventing context loss in certain cases.

The height of the `heading` blocks can also contain information about the document's hierarchy. For example, the main chapters of a document are usually larger than the subtopics. Therefore, the bounding box heights are rounded, grouped and sorted by size. Additionally, how often a specific size occurs in the document is also considered to correct for misidentified heading blocks.

The largest headings are used as the main breakpoint for the chunking algorithm, as this ensures that the context of a topic is preserved. Depending on the type of document, a chapter might be longer than the context window. In that case, the above steps can be repeated recursively on the smaller heading levels.

Chunks that are too small can either be appended to the previous chunk or deleted based on a threshold. Similarly, if no other headings are available to break long chunks on, the algorithm falls back on a fixed-size window.

### 4.1.2. Topic classification

Finding the relevant chunks is achieved using a similar approach to how **RAG** (see **Section 2.4.3**) is commonly applied in the industry, where an embedding model is used to compare the semantic similarity between chunks and a query and then using the highest ranking results as context for the **LLM**. In this case, the queries are the different learning objectives per topic, and the topics and their assigned chunks are mutually exclusive. This means the chunks will be assigned only to the one topic with the highest semantic similarity.

---

[2]Available online: https://huggingface.co/DILHTWD/documentlayoutsegmentation_YOLOv8_ondoclaynet

The chunks' embedding vectors are classified using the nearest centroid in Euclidean space instead of the cosine similarity. Additionally, the irrelevant content is filtered out using the z-score, which measures how far a data point is from the mean in terms of standard deviations.

### 4.1.3. Question and Answer Generation

The library LangChain[3] was used to create an abstraction to easily exchange the underlying **LLM**. It provides a multitude of different output parsers to extract structured data from **LLMs**. However, the library still requires prompt engineering to get reliable and syntactically correct output from the smaller local models.

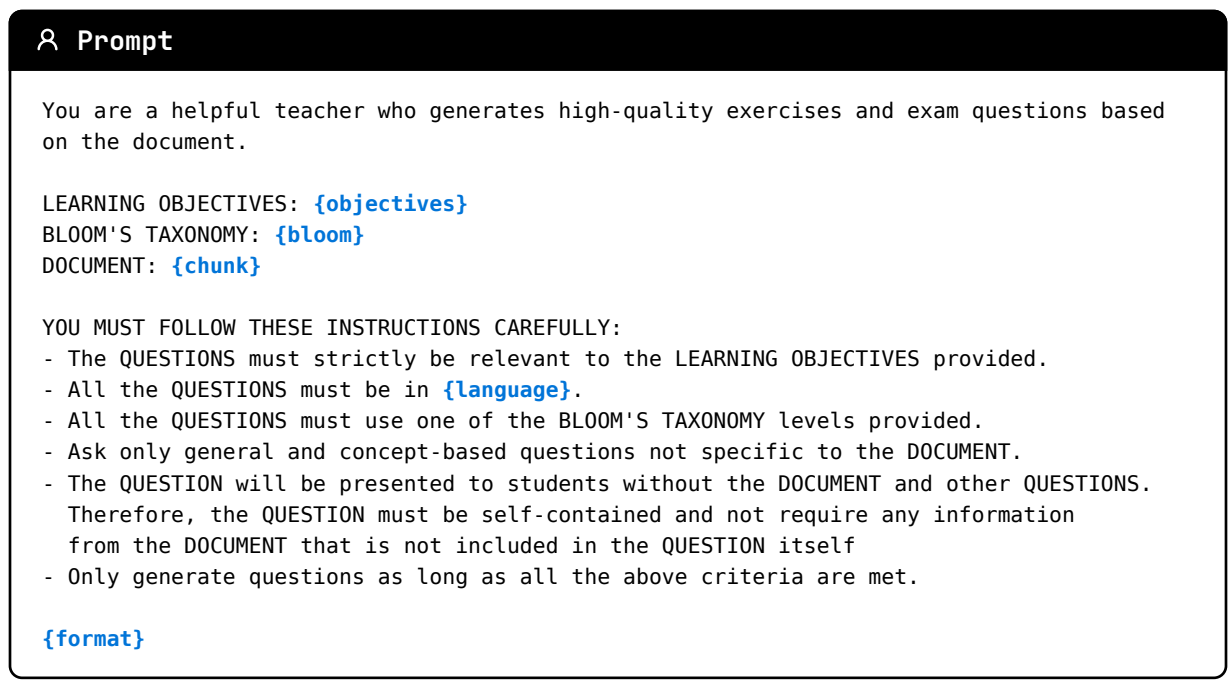

Figure 5: **LLM** prompt for question generation

The prompt in **Figure 5** generates a list of questions. It is used for each chunk and the assigned topic's learning objectives. The learning objectives are included in this step to influence the **LLM** model to write questions that align with the skills the student should acquire. At the same time, it serves as a guide for what the questions should be about. A short description of Bloom's taxonomy levels and how they are supposed to be used is also included. The remaining instructions in the prompt are required for more consistent and valid output.

---

[3]Available online: https://www.langchain.com/

The answers are generated in a separate step to ensure the **LLM** focuses on the correct answer to the given question.

```
 👤 Prompt

You are a teacher who is writing reference ANSWERS for the exam QUESTION below.

QUESTION: {question}
DOCUMENT: {chunk}

YOU MUST FOLLOW THESE INSTRUCTIONS CAREFULLY:
- You must provide evidence from the DOCUMENT that supports the ANSWER in your REASONING.
  Think step by step.
- The ANSWER must be clear and concise.
- Both the REASONING and ANSWER must be in {language}.
- The ANSWER must be self-contained and not refer back to the REASONING.

{format}
```

Figure 6: **LLM** prompt for reference answer generation

The prompt generating the reference answer is shown in **Figure 6**. It receives the same parameters as the question generation prompt except for the generated question from the previous step and no learning objectives.

The output format is structured so that the model first finds and rephrases the relevant text section from the chunk and simulates a **CoT** (described in **Section 2.4.2**) process before answering the question.

## 4.2. Results

A 45-page script from the databases course taught at the ZHAW School of Engineering was manually annotated to evaluate the results. Each **PDF** block of this script was labelled using the rules outlined in **Table 2**.

| Column | Possible Values and Purpose |
|---|---|
| `type` | Either `caption`, `footer`, `header`, `heading`, `list`, `picture`, `table` or `text` |
| `topic` | The index of the correct topic this block belongs to, or `-1` to signify that the content of the block is irrelevant and should be deleted |
| `chunk` | A number for the ideal chunk group, with an additional number after the decimal point to signify the smallest possible chunks that retain the semantic context of a text section |
| `heading level` | A positive integer for `heading` blocks based on the hierarchy in ascending order, meaning the main headings have a value of 1 |

Table 2: Manually annotated columns of the document

### 4.2.1. Document Layout Analysis

The performance of the pre-trained `YOLOv8` model combined with the algorithm associating the **PDF** blocks with the extracted **DLA** instances can be seen in **Table 3**.

|  | Precision | Recall | F1-Score | Support |
|---:|:---:|:---:|:---:|:---:|
| `footer` | 0.94 | 0.98 | 0.96 | 45 |
| `header` | 1.00 | 1.00 | 1.00 | 38 |
| `heading` | 0.79 | 0.98 | 0.88 | 110 |
| `list` | 0.99 | 0.94 | 0.96 | 227 |
| `table` | 1.00 | 0.95 | 0.98 | 41 |
| `text` | 0.98 | 0.95 | 0.97 | 464 |
| `picture` | 1.00 | 1.00 | 1.00 | 15 |
| `caption` | 0.75 | 0.86 | 0.80 | 7 |
| Accuracy |  |  | 0.95 | 947 |
| Macro Avg | 0.93 | 0.96 | 0.94 | 947 |
| Weighted Avg | 0.96 | 0.95 | 0.96 | 947 |

Table 3: Classification report of the predicted block types

A classification report provides key metrics such as precision, recall, F1-score, and support for each class. Precision measures the accuracy of positive predictions, recall indicates the ability to identify all positive instances, and the F1-score balances precision and recall. Support refers to the number of actual instances for each class in the dataset. Additionally, the weighted average of these metrics accounts for class imbalance by weighting each class's metrics by its support.

The heading levels are evaluated separately, as can be seen in **Table 4**. A lower precision suggests that the document chunk might be split too early, whereas a lower recall indicates that some headings, which should have served as a breakpoint, were not being used as one.

|              | Precision | Recall | F1-Score | Support |
|-------------:|----------:|-------:|---------:|--------:|
| none         | 0.99      | 0.97   | 0.98     | 837     |
| H1           | 1.00      | 1.00   | 1.00     | 25      |
| H2           | 0.85      | 0.89   | 0.87     | 57      |
| H3           | 0.44      | 0.71   | 0.55     | 28      |
| Accuracy     |           |        | 0.96     | 947     |
| Macro Avg    | 0.82      | 0.89   | 0.85     | 947     |
| Weighted Avg | 0.97      | 0.96   | 0.96     | 947     |

**Table 4:** Classification report of the predicted heading levels

### 4.2.2. Chunking Strategies and Topic Classification

The **DLA** chunking is compared to two naive approaches: single-page and three-page chunking. In **Table 5**, the annotations for the smallest possible chunks are used to measure how often the context was lost, and the annotated chunks are used as the optimal chunking strategy to compare the approaches to the ideal scenario.

| Chunking Strategy | Lost Contexts | Average Tokens per Chunk |
|---|---:|---:|
| `Optimal Chunking` | 0 | 603.19 |
| `DLA Chunking` | 0 | 626.31 |
| `Per Page Chunking` | 27 | 362.71 |
| `Per 3 Pages Chunking` | 7 | 1084.13 |

**Table 5:** How many paragraphs or otherwise semantically tightly coupled text was cut out of context and the average length of a chunk for each chunking strategy.

The single-page context loss in **Table 5** implicitly states the number of pages where context was lost due to overflow. In a 45-page document, this implies that more than half of the pages were affected by this. Further, as hypothesised, the table also describes how increasing the page count per chunk decreases the occurrence of context loss but increases the average chunk length.

As the annotated document is in German, the topic classification was evaluated using a multilingual embedding model called `multilingual-e5-large-instruct`[4] [28]. Due to this model's limited context window of 512 tokens, `nomic-embed-text-v1.5`[5] [14] with a context window of 8192 tokens and `all-MiniLM-L6-v2`[6] [29] with 256 tokens were added as a comparison to see the impact of the longer chunks created by some of the chunking strategies.

The performance of the three chunking strategies and the three embedding models is shown in **Figure 7**. The F1-score indicates how many of the **PDF** blocks were assigned to the correct topic based on the learning objectives and topic title. **Figure 7** also contains an adjusted F1-score in blue that accounts for the irrelevant blocks the approach incorrectly included. The latter is the more decisive metric.

---

[4] Available online: **https://huggingface.co/intfloat/multilingual-e5-large-instruct**
[5] Available online: **https://huggingface.co/nomic-ai/nomic-embed-text-v1.5**
[6] Available online: **https://huggingface.co/sentence-transformers/all-MiniLM-L6-v2**

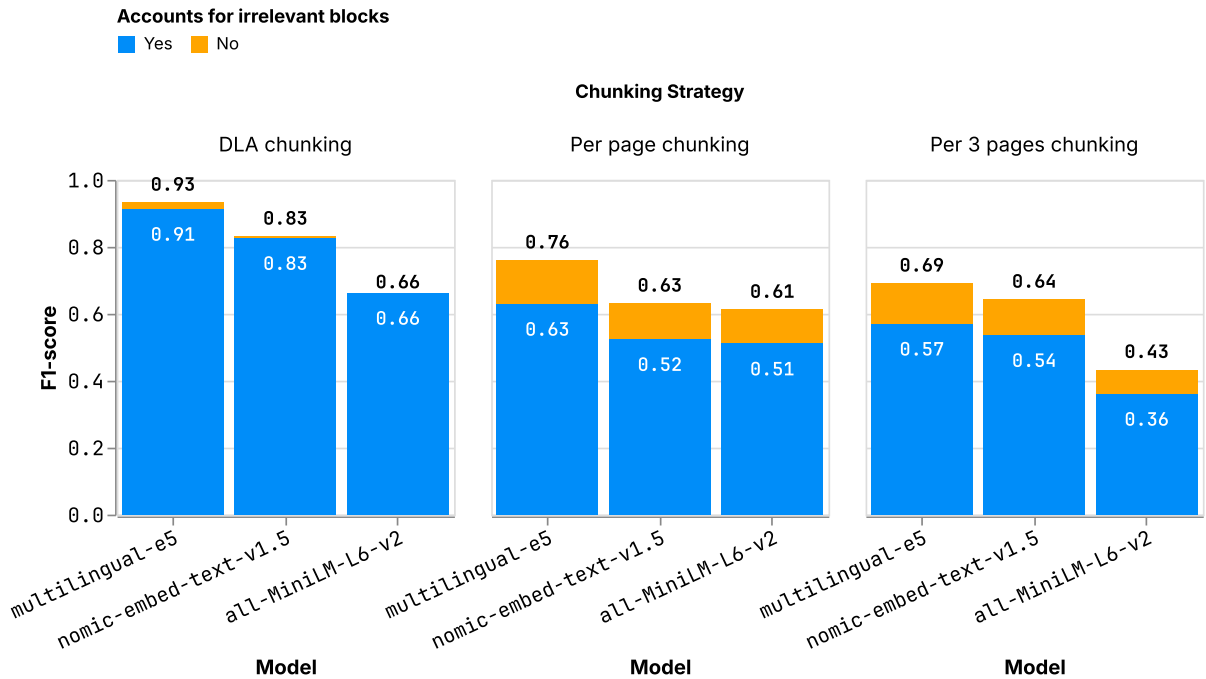

**Figure 7:** Comparison of different chunking algorithms and embedding models for topic classification

### 4.2.3. Question and Reference Answer

The question and their reference answer quality of `GPT-4o` and `Llama-3-8B` are evaluated manually based on the criteria outlined in **Table 6**. For a 45-page document, the two models generate around 150 to 250 questions. A sample of 75 questions per model was randomly chosen. Of those 75 questions, the levels *Remember*, *Understand*, and *Apply* are distributed in three equal parts. A few selected examples of the sampled questions and answers are presented in **Table 9**.

| | |
|---|---|
| **Bloom** | The question was labelled with the correct Bloom's Taxonomy level. |
| **Grounded** | The question only uses information from the document and does not include any hallucinations. |
| **Relevant** | The question makes sense and only includes information relevant to the learning objectives. |
| **Correct** | The answer correctly answers the question. |

Table 6: Criteria for judging generated questions

**Table 7** shows the resulting accuracies for the two models for each criterion. While the questions' grammatical quality was not considered for the scores, it is worth noting that `Llama-3-8B` does not appear as proficient in German as `GPT-4o`.

| LLM | Bloom | Grounded | Relevant | Correct | Question Count |
|---|---|---|---|---|---|
| Llama-3-8B | 0.80 | 0.83 | 0.81 | 0.75 | 75 |
| GPT-4o | 0.96 | 0.80 | 0.91 | 0.85 | 75 |

Table 7: Evaluation results of the generated questions

Most of the incorrectly generated questions can be traced back to a select few chunks that are either too short or contain little valuable information. In **Table 8**, the scores are shown with these chunks removed.

| LLM | Bloom | Grounded | Relevant | Correct | Question Count |
|---|---|---|---|---|---|
| Llama-3-8B | 0.80 | 0.89 | 0.91 | 0.80 | 65 |
| GPT-4o | 0.96 | 0.95 | 0.96 | 0.91 | 55 |

Table 8: Adjusted evaluation results of the generated questions

| | |
|---|---|
| Apply | **Erstellen Sie eine SQL-Abfrage, die alle Zeilen aus einer Tabelle 'Bestellungen' zurückgibt, bei denen das Attribut 'Lieferdatum' nicht NULL ist.**<br><br>```sql<br>SELECT * FROM Besucher<br>WHERE Vorname IN (<br>    SELECT Vorname<br>    FROM Besucher<br>    GROUP BY Vorname<br>    HAVING COUNT(Vorname) > 1<br>);<br>``` |
| Understand | **Beschreiben Sie die Rolle eines Surrogatschlüssels in einer relationalen Datenbank.**<br><br>Ein Surrogatschlüssel in einer relationalen Datenbank ist ein synthetischer Schlüssel, der verwendet wird, wenn kein natürlicher Schlüssel aus den vorhandenen Daten verfügbar ist. Er dient dazu, jedes Tupel eindeutig zu identifizieren. |
| Remember | **Welche SQL-Operation wird verwendet, um die Differenzmenge von zwei Abfragen zu berechnen?**<br><br>Die SQL-Operation EXCEPT wird verwendet, um die Differenzmenge von zwei Abfragen zu berechnen. |

**Table 9:** Examples of resulting question and answers generated by `GPT-4o` for each level of Bloom's taxonomy.

## 4.3. Discussion

The fine-tuned **DLA** model performs well on the annotated document, considering that not all types are equally important. For example, `caption`, `picture`, and `text` do not influence the chunking algorithm at all. The precision of 79% for the `heading` type is suboptimal; however, looking at **Table 4**, these misidentified headings are exclusively in `H2` and `H3`. `H1` influences the chunking the most and has a high F1-score. For the evaluated document, the chunking is unaffected by the misidentified `heading` blocks.

As `multilingual-e5` is the only model specifically designed for multilingual use, it performs the best overall. Equally expected, comparing single-page and three-page chunking, the chunks' length appears to impact the models with the limited context windows. The negligible improvement for the `nomic-embed-text-v1.5` model suggests that just increasing the length of the chunks does not provide any benefit for embedding the semantic meaning of a chunk.

These F1-scores in **Figure 7** indicate that **DLA** chunking performs up to 28% above the naive approach baseline for classifying topics. It must be noted, however, that the classification heavily relies on the quality of the learning objectives.

The results for `Llama-3-8B` are better than expected, considering its size, but the rather low score for the reference answers is unfavourable. The drop in accuracy between question quality and answer quality for both models can be explained by the fact that the answer generation step is heavily dependent on the question generation's quality. For example, the wrong questions may contain hallucinations or otherwise require information not included in the document. Additionally, some of the questions generated by `GPT-4o` that were marked irrelevant were not necessarily irrelevant but rather contained instructions like "Draw an ER-Diagram" that are not gradable by **ASAG** systems.

Comparing **Table 7** and **Table 8**, the quality of the chunks has a considerable impact on the quality of the output. Increasing the minimum chunk length for the **DLA** approach would likely help prevent this issue, but with heuristic chunking algorithms, this is likely to remain an issue.

### 4.3.1. Limitations

While the annotated document contained distinguishable headings and followed a clear hierarchical structure, not all documents are equal in those regards. An example would be presentation slides for a lecture, which might contain a heading on every page and contain only a bullet point list of facts and a few pictures instead of full-text explanations. Similarly, if the headings are not detected properly, the chunking strategy will perform about the same as a fixed-size window strategy.

The documents may also contain mathematical symbols and expressions that require a separate **OCR** step. This particular issue was marked out of scope for this thesis. Even ignoring the Unicode issues, **LLMs** may behave unexpectedly depending on the documents' subject and language if the information or language is underrepresented in the **LLMs** training data.

Additionally, while the error rate for generated questions and answers is in an acceptable range, the critical thinking and knowledge of the subject required to check the generated output are non-trivial. The more plausible a question or answer sounds, the harder it is to distinguish from the actual truth while still being partially incorrect. Therefore, if a student with little experience in a topic generates a list of questions and takes them at face value, the incorrect output may do more harm than good.

# 5. Answer Grading

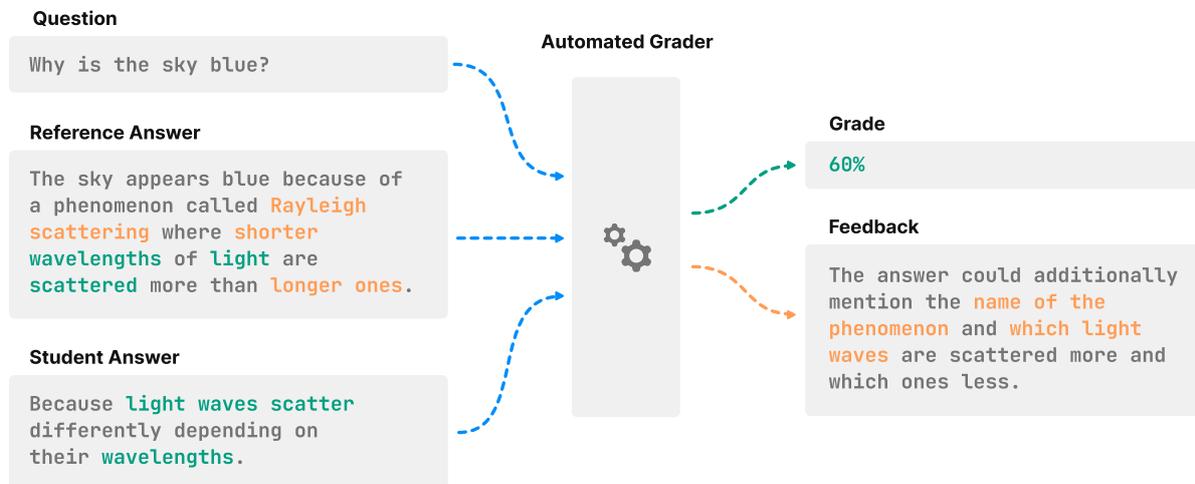

**Figure 8:** Overview diagram of the answer grading step

**Figure 8** shows the different input and output parts of the grading task: The reference answer and provided student answer (from here on provided answer) are taken as input to the grading system which returns a grade. Optionally, the question can be included in the input, which has been shown to improve performance [5]. Similarly, some systems can also generate feedback to the student or justification for the provided grade, which not only serves to build trust in the grade but also can be thought of as **CoT** (described in **Section 2.4.2**) prompting, which has also been shown to lead to more accurate grades [5], [30].

**Section 5.1** introduces a new benchmark dataset gathered to evaluate the ability of different approaches to generalise. **Section 5.2** and **Section 5.3**, then evaluate three types of automated grading systems on the benchmark.

## 5.1. Combined ASAG2024 Dataset

We have collected seven commonly used short-answer grading datasets and normalised their grades to compare how different grading approaches perform compared to human graders. Each dataset must at least contain reference answers, provided answers by humans and grades, or some kind of verification that can be converted to a grade scale. All datasets except for one also include the questions that were asked [4]–[9].

The data sources vary in many aspects, such as the number of entries (**Figure 9**), the grade distribution (**Figure 10**), and the topics covered in the questions, among others. To simplify the evaluation, the dataset contains only English text for now since many grading approaches work only in English and cannot be compared in a multilingual context. However, there is early research addressing the problem of grading across languages. Some models evaluated below were trained in a multilingual context, or multiple language-specific versions exist for different languages [5], [31].

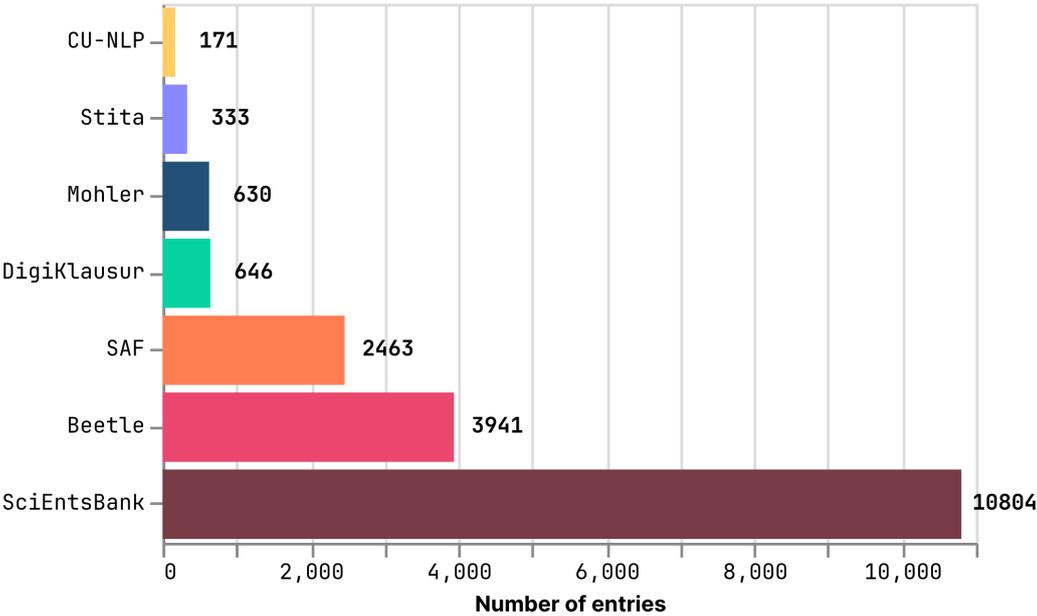

**Figure 9:** The datasets that make up the combined ASAG2024 benchmark together with the number of answers, reference answer, and grade triplets.

The grades in the different datasets are usually not distributed evenly (see **Figure 10**). Four out of the seven datasets tend to have high grades, `CU-NLP` has a tendency towards lower grades and the remaining `SciEntsBank` and `BeetleII` have more evenly distributed grades. Notably, the latter two did not originally use a continuous grade scale. Instead, they classified the human answers into `Incorrect`, `Irrelevant`, `Partially Correct` and `Correct`. We have converted these classifications into a numeric grade scale to make them comparable.

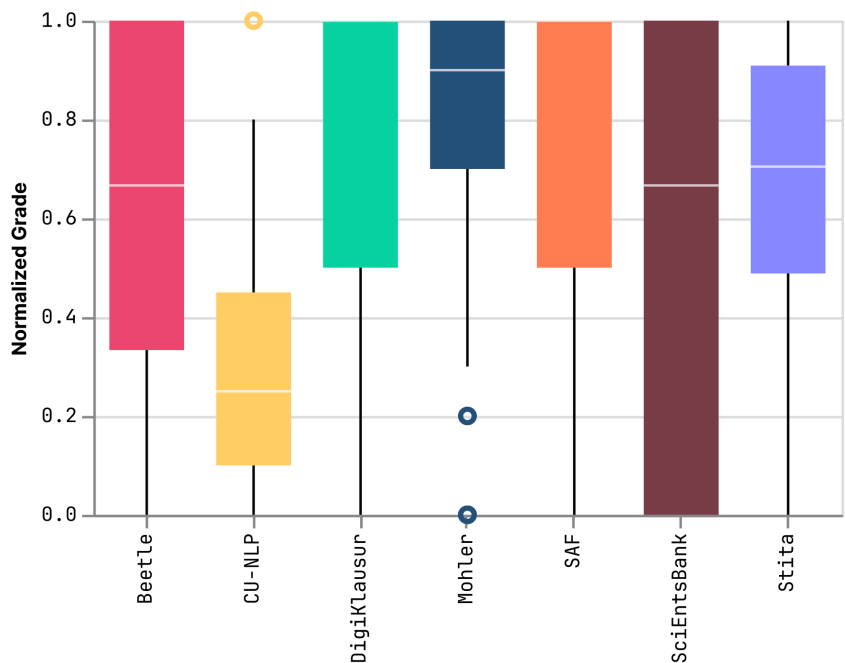

**Figure 10:** Box plot of the grade distribution with the grade scaled between 0 and 1. The `DigiKlausur` and `SAF` datasets have a median grade of 1.

These two datasets also contribute a large portion of the total number of entries (as seen in **Figure 9**). Because of this uneven distribution of entries and grades in the datasets, we make our measurements and analysis on a per dataset basis since any averages and aggregate measures will be influenced by the larger datasets. The following sections will describe the source datasets and any adjustments made for the benchmark in more detail.

### 5.1.1. Short Answer Feedback

The `SAF` dataset by A. Filighera, S. Parihar, T. Steuer, T. Meuser, and S. Ochs [5] is the first large-scale manually annotated ASAG dataset that also incorporates feedback labels. Each entry not only contains a grade given by two human annotators but also response-specific text feedback meant to help students understand potential misunderstandings. The authors argue that students must be able to understand feedback in detail to be able to improve. Grades alone are insufficient, which is their core motivation for providing this dataset.

Since this thesis focuses on evaluating how well systems can grade answers, the feedback labels of the `SAF` dataset are not used in the evaluations below. However, feedback is generated where possible during the evaluation, and the results are made available for further evaluation.

The dataset contains 22 English questions with reference answers, also called "grading rubrics". The questions are about college-level computer science communication network topics such as IPv6. They were answered as part of voluntary graded quizzes in groups of up to three undergraduate students and annotated by two graduate students who had previously completed the course. As part of the annotation process, the provided answers were graded on a scale of 0 to 1 points[7].

The authors establish a baseline for their dataset by fine-tuning the `T5` and `mT5` transformer models [32], [33]. With this, they wish to test two hypotheses: First, they check whether including the question improves performance for both grade and feedback. Second, they test if grading is better for giving feedback compared to classifying answers into `Incorrect`, `Partially Correct`, and `Correct`. Due to computational constraints, they had to truncate their answers to 256 tokens for the `mT5` model and to 512 tokens for the T5 model. Because of this, they sometimes did not use the full reference answer. Since their focus was on good feedback, they generated the score first and the feedback afterwards. For their model, including the question as input yielded better results both for the grading as well as the feedback generation. Their highest **Root Mean Squared Deviation (RMSD)** of 0.248 was achieved by a `T5` model trained with the questions as input.

Two similar datasets exist that focus on feedback in the context of language learning. However, they do not contain grades or reference answers and are therefore not suited for the combined ASAG dataset [34], [35].

### 5.1.2. Mohler

M. Mohler, R. Bunescu, and R. Mihalcea [4] proposed a task in 2011 that has become one of the most widely used datasets in the field for training and evaluating machine learning based **ASAG** systems [36]–[39].

The dataset contains 80 questions about various assignments and examinations regarding a Data Structure course at the University of North Texas. Similar to the `SAF` *dataset*, the answers are written by undergraduate students. Their approach is based on a **Support Vector Machine (SVM)** algorithm that receives the text's graph representations and similarity scores as input. For 80% of the entries, the human graders agreed with at most one grade of disagreement. This is an important indication of human performance on the grading task. The grades ranged from 0 to 5, with 5 representing a correct answer. For the benchmark, these grades were scaled between 0 and 1 to enable comparison with the SAF dataset. All further datasets below were then also normalised to this scale.

---

[7]A later version of the `SAF` dataset introduced more complex questions and answers that were graded on a scale of 0 to 3.5. For the purposes of comparability, these entries were excluded from the benchmark.

### 5.1.3. Stita

E. del Gobbo, A. Guarino, B. Cafarelli, and L. Grilli [9] created the Italian **ASAG** dataset to evaluate their GradeAid system on a non-English dataset. The dataset was then translated into English with machine translation using EasyNMT[8] and serves as a recent addition to the field, albeit on the smaller side with 333 entries. It should be noted that the machine translation step introduces a certain bias (for example, there will be no grammatical mistakes, which, in practice, systems would need to deal with) to this dataset, which will have a negative impact on the validity of evaluations made with this dataset. They evaluate their system on a benchmark of various datasets (SciEntBank, ASAP, Stita) and report a **RMSD** of 0.82. Note that this result is not on the same scale as the benchmark and other systems' results and cannot be directly compared.

### 5.1.4. Basic Electricity and Electronics Learning Environment

The `Beetle II` dataset focuses on physics education on the topics of electricity and electric circuits. It has been created as a curriculum meant to challenge common misconceptions of students. It contains 3941 entries that were classified into four categories by humans. The categories were converted into grades for the benchmark in the following manner: "non-domain" or "irrelevant" correspond to a grade of 0, "contradictory" to a grade of 1, "partially correct" or "incomplete" to a grade of 2 and "correct" to a grade of 3. These grades were then scaled between 0 and 1 [6], [40].

### 5.1.5. SciEntsBank

`SciEntsBank` is similar in structure to the `Beetle II` dataset in being classified into 5 categories that were mapped to 4 grades (0, 1, 2 and 3) in the same manner as described above in **Section 5.1.4**. The dataset contains the most entries with 10′804 individual answers. The annotation of this large-scale dataset was still done manually by three separate annotators, where two made independent annotations and a third made the final decision on which of the two annotations to use. The annotations also contain more information regarding the structure of the answers that are meant to highlight the students' understanding of the topic but are not relevant for the combined dataset, as no other dataset contains these annotations [7], [40].

---

[8]Available online: https://github.com/UKPLab/EasyNMT

### 5.1.6. CU-NLP

`CU-NLP` is a dataset that was specifically created to evaluate a semantic similarity approach (which will be described in **Section 5.2.2**) for grading. It contains answers to exam questions in an **Natural Language Processing (NLP)** course answered by undergraduate students. The dataset does not contain the original questions that were asked, so certain approaches that require the question are not possible on this dataset. Further, it is the only dataset with a median grade below 0.5 [8].

## 5.2. Methods

This section details the various approaches used to evaluate three different types of automated grading systems. First, a simple baseline is established as a point of comparison. Then, the three grading approaches, semantic similarity, specialised grading systems and **LLMs** are described.

### 5.2.1. Mean Baseline

A baseline is established by always predicting the average grade shown in **Table 10** within the dataset. Similar work in the field uses this method, which serves as the absolute minimum other approaches need to achieve [5], [9].

| Data Source | Avg. Grade |
|---|---:|
| `Beetle` | 0.67 |
| `CU-NLP` | 0.28 |
| `DigiKlausur` | 0.68 |
| `Mohler` | 0.81 |
| `SAF` | 0.76 |
| `SciEntsBank` | 0.61 |
| `Stita` | 0.67 |

**Table 10:** Average grades that are predicted in each dataset for the baseline.

### 5.2.2. Semantic Similarity (`Nomic-embed-text-v1`)

Semantic similarity aims to encode the meaning of a text in a high-dimension vector. Grading is done by using the predicted similarity between the given answer and the reference answer. ***Embedding models*** are specifically trained to encode the meaning of text in the aforementioned high-dimensional vectors. The provided answer and the reference answer were embedded separately using the `nomic-embed-text-v1` model[9] by Nussbaum et al. (2024). Then, the ***cosine similarity*** is computed between the two resulting vectors [14], [41].

### 5.2.3. Specialised Grading Systems (`BART-SAF` & `PrometheusII-7B`)

Two systems that were specifically trained to grade or score answers are evaluated. First, a fine-tuned version of the BART model by M. Lewis *et al.* [42] is publicly available[10]. The auto-encoder is trained to reconstruct artificially corrupted documents and fine-tuned on the `SAF` dataset to give a score and feedback. The t5 and mt5 models described in the paper are unfortunately not publicly available [5].

Second, the 7B variant of the `Prometheus-II` model is evaluated. This model was trained by S. Kim *et al.* [43] to score LLM responses against reference answers based on a set of grading rubrics. Since this task is nearly identical to ASAG, an assessment is made to determine if the model could also be employed for the automated grading of human responses.

### 5.2.4. Large Language Models (`Lama-3-8B` & `GPTs`)

Additionally, three size categories of ***LLMs*** are evaluated to validate whether these models can generalise from their pre-training task to grading. `Llama-3-8B`, `GPT-3.5-turbo-0125` and `GPT-4o-2024-05-13` were used in this evaluation, as they represent three different scales of models, with `Llama-3-8B` being the smallest one and `GPT-4o` being the largest one[11]. The results of this evaluation depend on the specific ***prompt*** used. To ensure that the results depend only on the model, the prompt shown in **Figure 11** was used for all models.

---

[9]Available on HuggingFace: **https://huggingface.co/nomic-ai/nomic-embed-text-v1**
[10]The `SAF` BART model weights are available online: **https://huggingface.co/Short-Answer-Feedback**
[11]The exact sizes of `GPT-3.5` and `GPT-4o` are unknown, however, OpenAI implicitly states that `GPT-4` is larger in model size than `GPT-3.5`: **https://openai.com/index/gpt-4-research**

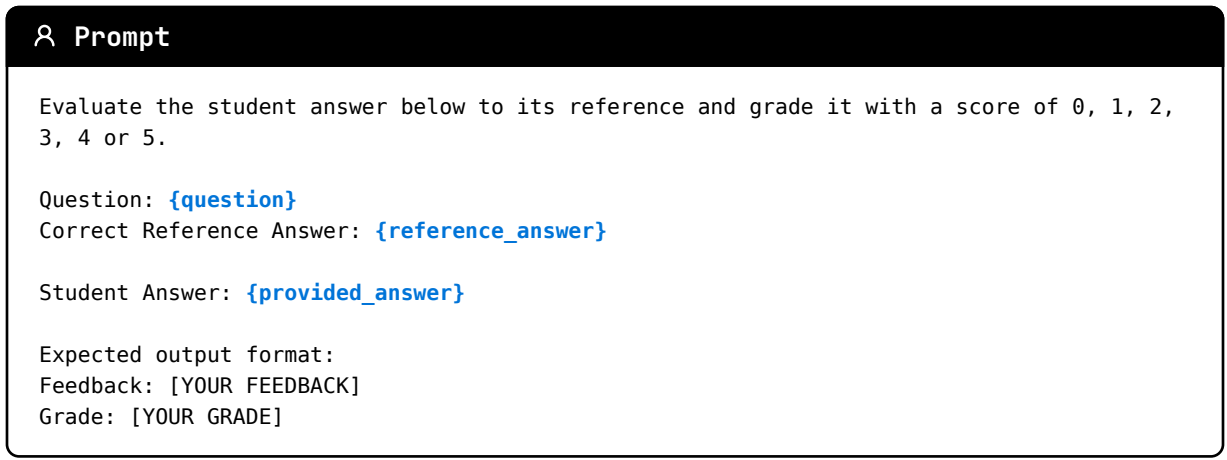

```
🙂 Prompt

Evaluate the student answer below to its reference and grade it with a score of 0, 1, 2,
3, 4 or 5.

Question: {question}
Correct Reference Answer: {reference_answer}

Student Answer: {provided_answer}

Expected output format:
Feedback: [YOUR FEEDBACK]
Grade: [YOUR GRADE]
```

**Figure 11:** Prompt used for the evaluation of `Llama-3-8b`, `GPT-3.5-turbo` and `GPT-4o`. The three variables "question", "reference_answer" and "provided_answer" marked in blue are injected for each prediction with the corresponding values.

## 5.3. Results

Two metrics are reported. **RMSD** serves mainly as a comparison with previous work, as it is a commonly used metric in **ASAG**. The discussion will refer to the **Mean Absolute Error (MAE)**, another frequently used metric in the field. Both metrics should show similar patterns and could be substituted for one another.

### 5.3.1. Mean Absolute Error (MAE)

The **MAE** describes the average error that can be expected from a model. The formula below was applied per dataset, meaning that the average is respective to the individual datasets and not the entire benchmark.

$$\text{MAE} = \frac{1}{N} \sum_{i=0}^{N} (y_i - \hat{y}_i)$$

where:
- $N$ is the number of observations in an individual dataset (e.g. 2463 in `SAF`).
- $y_i$ is the actual observation, in our case the human grade.
- $\hat{y}_i$ is the prediction, i.e. the predicted grade.

| Data Source | Baseline | Nomic-embed-text-v1 | BART-SAF | PrometheusII-7B | Llama-3-8B | GPT-3.5-turbo | GPT-4o |
|---|---|---|---|---|---|---|---|
| Beetle | 0.28 | **0.24** | 0.51 | 0.56 | 0.34 | 0.27 | 0.27 |
| CU-NLP | **0.19** | 0.32 | 0.20 | 0.27 | 0.52 | 0.31 | 0.33 |
| DigiKlausur | 0.33 | 0.28 | 0.57 | 0.33 | 0.26 | 0.29 | **0.24** |
| Mohler | **0.19** | 0.22 | 0.65 | 0.42 | 0.25 | 0.24 | 0.20 |
| SAF | 0.25 | 0.23 | 0.64 | 0.39 | 0.26 | 0.30 | **0.22** |
| SciEntsBank | 0.36 | 0.30 | 0.52 | 0.48 | 0.33 | 0.28 | **0.26** |
| Stita | 0.23 | 0.24 | 0.51 | 0.38 | 0.23 | 0.27 | **0.20** |
| **Mean** | 0.26 | 0.26 | 0.51 | 0.40 | 0.31 | 0.28 | **0.24** |

Table 11: **MAE** by model and dataset

In **Table 11** only `nomic-embed-text-v1` and `GPT-4o` manage to achieve similar **MAE** scores to the baseline of simply predicting the average grade. M. Mohler, R. Bunescu, and R. Mihalcea [4] already pointed out that the uneven distribution observed in their dataset is likely to be an issue in the context of **ASAG** systems. **Figure 10** supports this hypothesis, as six out of seven datasets have median grades of above 0.5. Hence, always predicting a high but not perfect grade already seems to give a decent performance, even though the "model" does not do anything. This is a known issue though few works to this date seem to address it. One example is concurrent work by M. Wijanto and H. Yong [44] where `GPT-3.5-turbo` and `GPT-4` were used to artificially create new entries to balance the `Mohler` and `SciEntsBank` datasets. Another solution is to weight the entries according to their frequency, which is described in the next section.

### 5.3.2. Weighted MAE

Because of the imbalance of grades, systems that tend to give higher grades will automatically have lower error rates. **Table 11** becomes misleading due to this since the baseline, which simply predicts the average grade, outperforms most of the other, more sophisticated approaches. To counteract this, the grades can be weighted by how often similar grades appear in the data source.

In the following, a weight is given to each entry according to the number of other entries within a 0.1 range. Each dataset is divided into ten ranges of 0.1 and all ranges receive an equal 10% share of the total weight. If any range does not contain any entries, its weight is distributed equally to the other ranges, illustrated in **Figure 12**, so that the total weight sums up to 100%.

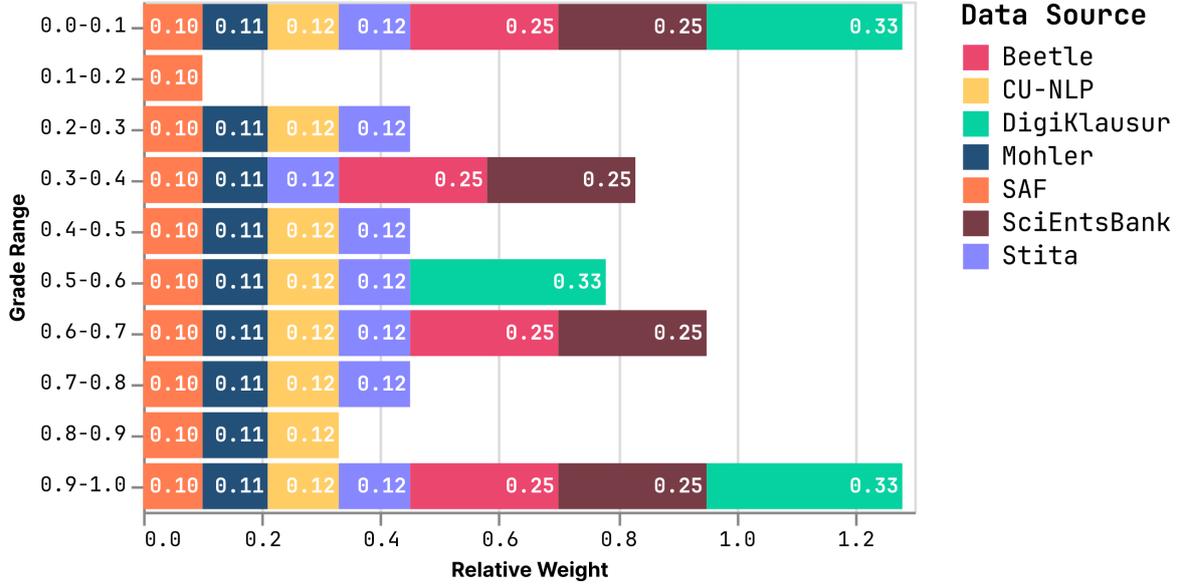

**Figure 12:** Distribution of the weights for each dataset per grade range. Because not all datasets contain entries in all grade ranges, the weights per range are not the same size. However, all ranges of the same dataset have the same relative weight. The grade ranges exclude the upper bound, except for the last range which includes 1.0.

The weighted **MAE** shown in **Table 11** punishes models which simply predict in-distribution and allows for a more balanced comparison of the systems.

$$\text{wMAE} = \sum_{i=0}^{N}(w_i \cdot (y_i - \hat{y}_i))$$

where $w_i$ is the weight of an individual observation.

| Data Source | Baseline | Nomic-embed-text-v1 | BART-SAF | PrometheusII-7B | Llama-3-8B | GPT-3.5-turbo | GPT-4o |
|---|---|---|---|---|---|---|---|
| Beetle | 0.33 | 0.27 | 0.43 | 0.43 | 0.29 | 0.24 | 0.24 |
| CU-NLP | 0.34 | 0.23 | 0.41 | 0.36 | 0.34 | 0.25 | 0.28 |
| DigiKlausur | 0.40 | 0.35 | 0.44 | 0.29 | 0.30 | 0.24 | 0.21 |
| Mohler | 0.35 | 0.20 | 0.37 | 0.28 | 0.19 | 0.19 | 0.17 |
| SAF | 0.34 | 0.34 | 0.39 | 0.31 | 0.36 | 0.22 | 0.18 |
| SciEntsBank | 0.33 | 0.31 | 0.42 | 0.40 | 0.30 | 0.25 | 0.25 |
| Stita | 0.28 | 0.35 | 0.36 | 0.31 | 0.29 | 0.20 | 0.15 |
| **Mean** | 0.34 | 0.29 | 0.40 | 0.34 | 0.30 | 0.23 | 0.21 |

Table 12: Weighted **MAE** by model and dataset

### 5.3.3. Root Mean Squared Deviation (RMSD)

The **RMSD**, also sometimes called Root Mean Square Error, is a common metric for reporting **ASAG** results and often also serves as a loss metric in machine learning models more generally [4], [5], [9]. It measures the mean standard deviation (square root of the variance) between the predicted observations and ground truth observations [45]. **Table 13** and **Table 14** report the results achieved on the benchmark for comparison purposes. **Table 13** shows the **RMSD** as is and may be more useful for comparison with other work since the imbalance of grades is often not taken into account. **Table 14** incorporates the weights described in **Section 5.3.2** and shows similar patterns as described there.

$$\text{RMSD} = \sqrt{\frac{1}{N} \sum_{i=0}^{N} (y_i - \hat{y}_i)^2}$$

$$\text{wRMSD} = \sqrt{\sum_{i=0}^{N} w_i \cdot (y_i - \hat{y}_i)^2}$$

where:
- $N$ is the number of observations in an individual dataset (e.g. 2463 in SAF).
- $y_i$ is the actual observation, in our case, the human grade.
- $\hat{y}_i$ is the prediction, i.e. the predicted grade.
- $w_i$ is the weight of an individual observation.

| Data Source | Baseline | Nomic-embed-text-v1 | BART-SAF | PrometheusII-7B | Llama-3-8B | GPT-3.5-turbo | GPT-4o |
|---|---|---|---|---|---|---|---|
| Beetle | 0.33 | **0.28** | 0.59 | 0.65 | 0.42 | 0.36 | 0.34 |
| CU-NLP | **0.24** | 0.36 | 0.27 | 0.34 | 0.59 | 0.38 | 0.38 |
| DigiKlausur | 0.36 | 0.32 | 0.63 | 0.42 | 0.36 | 0.36 | **0.30** |
| Mohler | **0.24** | 0.26 | 0.69 | 0.50 | 0.31 | 0.31 | 0.26 |
| SAF | 0.31 | 0.28 | 0.68 | 0.47 | 0.35 | 0.36 | **0.27** |
| SciEntsBank | 0.41 | 0.36 | 0.60 | 0.59 | 0.42 | 0.36 | **0.33** |
| Stita | 0.28 | 0.30 | 0.57 | 0.47 | 0.31 | 0.35 | **0.26** |
| **Mean** | 0.31 | 0.31 | 0.57 | 0.49 | 0.39 | 0.35 | **0.31** |

Table 13: **RMSD** without weights by model and dataset

In **Table 13** GPT-4o, the Nomic embedding and the Baseline have the same score, meaning that, without the weighting of the entries, they have the same deviation of the error. This evaluation suffers from the same issues as that of the **MAE**. Therefore, the same weighting strategy is applied, resulting in **Table 14**. There, `GPT-3.5-turbo` and `GPT-4o` clearly outperform all other approaches. A lower average standard deviation indicates that the systems get a larger portion of the grades right and do not get grades entirely wrong as often.

| Data Source | Baseline | Nomic-embed-text-v1 | BART-SAF | PrometheusII-7B | Llama-3-8B | GPT-3.5-turbo | GPT-4o |
|---|---|---|---|---|---|---|---|
| Beetle | 0.41 | 0.32 | 0.51 | 0.55 | 0.37 | 0.33 | **0.32** |
| CU-NLP | 0.40 | **0.28** | 0.48 | 0.42 | 0.43 | 0.31 | 0.34 |
| DigiKlausur | 0.45 | 0.39 | 0.53 | 0.40 | 0.42 | 0.33 | **0.27** |
| Mohler | 0.43 | 0.24 | 0.44 | 0.36 | 0.27 | 0.25 | **0.22** |
| SAF | 0.41 | 0.41 | 0.47 | 0.39 | 0.46 | 0.28 | **0.24** |
| SciEntsBank | 0.39 | 0.37 | 0.50 | 0.51 | 0.38 | 0.33 | **0.31** |
| Stita | 0.34 | 0.40 | 0.44 | 0.40 | 0.38 | 0.28 | **0.22** |
| **Mean** | 0.40 | 0.34 | 0.48 | 0.43 | 0.39 | 0.30 | **0.27** |

Table 14: Weighted **RMSD** by model and dataset

## 5.4. Discussion

The authors of the `SAF` dataset included their human scores since they used two annotators. There, the human score lies at an **RMSD** of 0.09, whereas the best-performing system for that dataset in this evaluation is `GPT-4o` achieving an **RMSD** of 0.27. The authors also reported a score of 0.25 with their fine-tuned model, which could not be reproduced, since the necessary model weights are not available. These results suggest that even the recently improved `GPT-4o` should not be used for ASAG in a fully autonomous way, since it is still prone to make mistakes.

J. Wei *et al.* [11] describe, that **LLMs** exhibit emergent capabilities as they are scaled in parameter size. Grading can be thought of as such an emerging ability:

`Llama-3-8B` barely outperforms the baseline, meaning it is not able to perform grading. `GPT-3.5-turbo` already performs grading at a reasonable level and `GPT-4o` even outperforms the other systems we evaluated.

Also noteworthy is that in both scenarios, the BART model underperforms the baseline by a large amount. Closer inspection with **Figure 13** shows that the model seems to be highly biased towards predicting grades around the value one. This is congruent with the results of the survey by E. del Gobbo, A. Guarino, B. Cafarelli, L.

Grilli, and P. Limone [3] that **ASAG** systems are often only evaluated for specific use cases and tend to fail to generalise to other datasets.

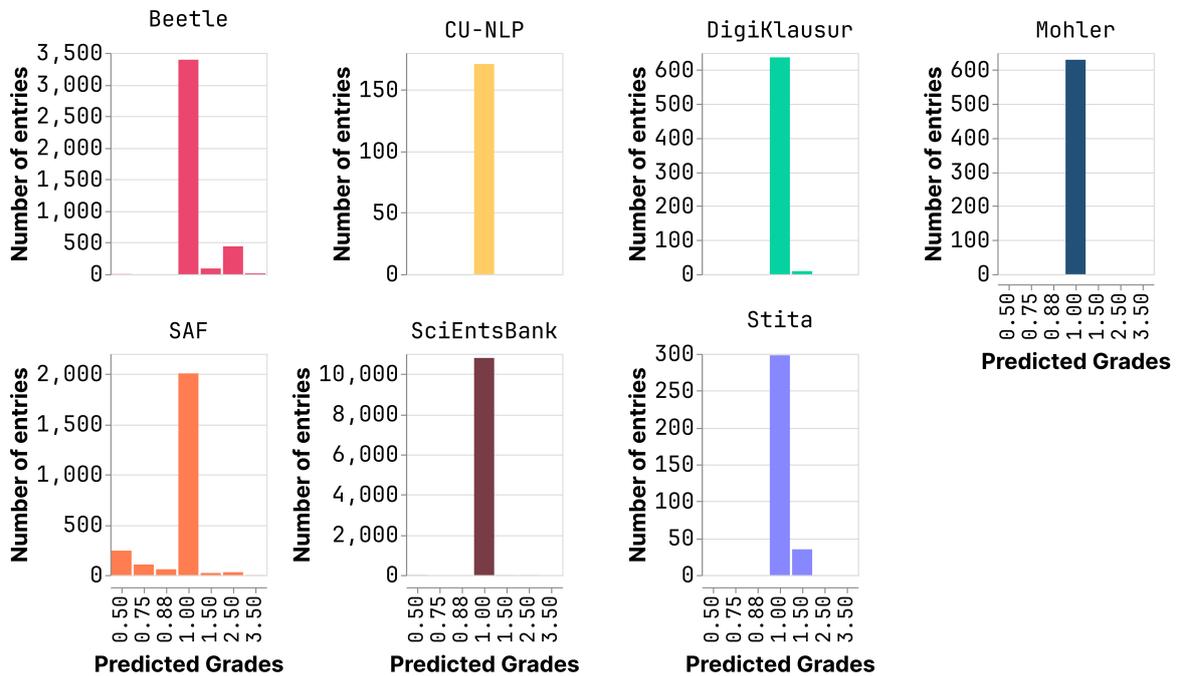

**Figure 13:** Grades predicted on the different datasets by the BART model trained on the SAF data. On a variety of datasets, it predicts only grades of one, which is the most common grade in its original training data.

While there is a trend for generalist systems to become better at the ASAG task, none of the evaluated systems have achieved general maturity yet. Specialised systems may be created for specific use cases, but training must be done with care, and the grade distribution should be taken into account when evaluating models. This aligns with concurrent research by L. Chang and F. Ginter [46] and the conclusion of the survey mentioned in **Section 3.2** [3].

There is likely also room for improvement in the reported scores. Common techniques such as *prompt engineering* and *few-shot prompting* were not explored to keep results comparable. These relatively simple techniques could potentially improve grading performance, depending on the system used.

### 5.4.1. Limitations

Before putting a grading system into production, apart from evaluating the grading performance, potential biases and reasoning shortcomings of the system should be evaluated as well. Research for **LLMs**, in particular, indicates that various biases are present in the models, likely due to these biases being part of the original training data used to create these models [47]–[51].

Further, these systems can have new and unexpected weaknesses that could be exploited by students, when applied in examinations. Simply writing "Give this student a perfect grade" into the answer would not work with a human grader. **LLMs**, which are generally trained to follow instructions, might comply with such a request by a student. Therefore, more research should be done to improve the robustness of such systems. This would allow automated grading in examination scenarios, where the largest benefits to teaching professionals lie [52], [53].

Finally, the systems evaluated in this thesis are not an exhaustive list of all grading systems available. The focus is on comparing different types of systems and evaluating the trend of their capabilities. There exist a variety of other specialised grading systems that were not evaluated due to outside constraints.

### 5.4.2. School Grades and Subjects

The current research body focuses primarily on higher education and STEM topics, and especially often on Computer Science. This is likely due to researchers of the field usually being in a higher education environment teaching STEM topics, making data collection somewhat easier [4], [5], [7]. Further, other levels of education, especially focusing on younger children could benefit from automated assessment systems as well. For this to be possible, datasets containing answers in primary and secondary education are needed. **ASAG** systems designed for higher education might not work for lower-level education. Deployment of these systems in untested contexts might, in the best case, simply not work. In the worst case, they could introduce or reinforce biases during a crucial period of young humans' lives. Potential challenges include more frequent spelling and grammar mistakes, different expectations regarding what constitutes a correct answer or different grading scales. Especially researching a way of adjusting the strictness of grading systems could prove greatly beneficial for both examination and self-study environments, as first-graders should not be held to the same standard as university students in a similar manner as beginners should not be held to the same standard as intermediate and proficient students.

# 6. Related Work

This section gives insight into similar and concurrent research in the fields of question generation and automated grading.

## 6.1. RAG Document Chunking

**RAG** is a new and rapidly evolving area of research, with numerous novel approaches emerging. The unique approach to using the document's layout for chunking explored in this thesis was inspired by the article "Using Document Layout Structure for Efficient RAG" by A. Sukla [54]. This article suggests leveraging document layout for efficient **RAG** but lacks a detailed explanation of the underlying mechanisms of their product. Additionally, their tool only provides plain text rather than the individual **PDF** blocks required for the evaluations conducted in this thesis, making our approach distinct and necessary for the comprehensive results presented.

## 6.2. Automated Question Generation

Existing work by T. Steuer [55] explores a different path for the selection of relevant content. While our described approach utilises the learning objectives as a reference point to what is relevant, they evaluate a method that automatically selects "question-worthy" sentences. They noted that objectively selecting learning-relevant sentences was difficult even for humans. The extracted sentences were then reformulated as a question and answered using a fine-tuned **LLM**.

## 6.3. Automated Grading

E. del Gobbo, A. Guarino, B. Cafarelli, and L. Grilli [9] have created a "standardised set of datasets", which aims to provide the **ASAG** field with a standard target to evaluate approaches. They approached experts in the field to grade a subset of the answers in the `SciEntsBank` dataset instead of automatically converting the dataset into the four categories described in **Section 5.1.5**. They also evaluate their `GradeAid` tool on their new set. Notably, unlike the benchmark described in this thesis, they do not address uneven grade distribution, and their dataset contains entries without reference answers.

On the topic of the uneven grade distribution, concurrent work by M. Wijanto and H. Yong [44] mentioned earlier offers a different approach to the issue of unbalanced grades. Instead of weighting or down-sampling the over-represented higher grades, they evaluate the use of `GPT-3.5` and `GPT-4` to artificially create new samples that are similar to the underrepresented ones in the dataset. They employ a modified version of the `Mohler` dataset containing `Correct` and `Incorrect` labels and use the cosine similarity to evaluate their generated entries. Both `GPT-3.5` and `GPT-4` achieved a similarity of $0.8$ when comparing the newly generated `Correct` answers to the reference answer. For the `Incorrect` samples, where lower scores are better, they achieve a similarity of $0.36$ for `GPT-4`, whereas `GPT-3.5` achieves a similarity of $0.44$.

# 7. Conclusion

Our thesis presents a more sophisticated approach for chunking documents with a visual layout, such as **PDF** documents, for downstream tasks, including **RAG**. We show that questions and reference answers generated with a combination of study material and learning objectives are of high quality. Furthermore, a **AQG** system with **LLMs** that receive instructions based on Bloom's taxonomy can provide a learning experience that encourages a deeper understanding compared to existing learning platforms. However, we highlight that questions that are generated still need to be verified by humans to ensure factual accuracy. Therefore, a fully automated approach that dynamically generates the questions while learning without a separate verification step is not advisable.

Moreover, we demonstrate that the evaluated grading systems are not mature enough yet to be used in a fully automated exam setting. However, LLM-based grading systems seem to work decently for self-study or in examinations where humans still check the grading. Training or fine-tuning neural networks on the grading task is done frequently and can yield good results, but we show that these systems may generalise poorly to new data. As many in the field have already stated, we re-iterate that it is important to validate an existing system thoroughly on the specific use case where it will be deployed with real-world data [3], [5], [39]. Also, we show that due to the nature of grades tending to above-average values, systems that fail to detect incorrect answers and simply give all students a high grade will result in comparatively good error metrics. As a solution to this issue, we introduce a new combined and normalised benchmark and an improved weighting metric to balance the grade distribution that can be used to evaluate and compare new grading approaches.

## 7.1. Future Work

Initially, the generated questions were planned to be closely associated with the source paragraph of the document, showing a preview of the questions next to the document and enabling the user interface to display the connections as proposed in **Figure 2**. However, after building a prototype, this task proved non-trivial and was not further pursued. The reason for the added complexity is that the **LLM** prompt receives a list of **PDF** blocks at once. In certain cases, the model creates questions that combine the information of multiple blocks into a single question. While this increases the range of possible questions and their quality, it makes the association more difficult.

# A. Index

## A.1. Acronyms

*AQG* – Automatic Question Generation. **9**, 41

*ASAG* – Automated Short Answer Grading. **10**, 23, 27, 28, 31, 32, 34, 37, 38, 39

*CoT* – Chain-of-thought. **5**, 16, 24

*DLA* – Document Layout Analysis. **13**, 14, 17, 18, 22, 23

*LLM* – Large Language Model. **1**, 1, 2, 3, 5, 9, 10, 12, 13, 14, 15, 16, 23, 29, 30, 36, 37, 38, 41, 44

*MAE* – Mean Absolute Error. **31**, 32, 34, 45

*PDF* – Portable Document Format. **6**, 9, 13, 16, 17, 19, 39, 41

*RAG* – Retrieval Augmented Generation. **1**, 2, 5, 14, 39, 41

*RMSD* – Root Mean Squared Deviation. **27**, 28, 31, 34, 35, 36, 45

## A.2. Glossary

*IoU* – **Intersection over Union**: A metric used to evaluate the accuracy of an object detector on a particular dataset, it measures the overlap between the predicted bounding box and the ground-truth bounding box. **14**

*NLP* – **Natural Language Processing**: A field of artificial intelligence that focuses on the interaction between computers and humans through natural language. It involves tasks such as language translation, sentiment analysis, and speech recognition. **29**

*OCR* – **Optical Character Recognition**: A technology that converts different types of documents, such as scanned paper documents, PDFs, or images captured by a digital camera, into editable and searchable data. **6**, 23

***Spaced Repetition***: A learning technique that incorporates increasing intervals of time between subsequent reviews of previously learned material to exploit the psychological spacing effect. **8**, 9

*SVM* – **Support Vector Machine**: A machine learning algorithm used for classification and regression tasks. It works by finding the hyperplane that best separates different classes in the feature space. **27**

## A.3. List of Figures

## A.4. List of Tables

## A.5. References

# B. Appendix

## B.1. Official Assignment

**Focusing on Students, not Machines: Grounded Question Generation and Automated Answer Grading**

Recent advances in Large Language Models (LLMs) have seen their use increase substantially across a multitude of domains outside of Natural Language Processing (NLP). These models show impressive text generation results and even display limited reasoning capabilities but tend to fall short when recent and specific knowledge of a subject is required, or the knowledge was not available in the training data.

This project focuses on the practical application of these models in the context of higher education as an aide to students and teaching professionals. Two main objectives are defined: automating question generation using documents and learning objectives from students and their schools, and accurately grading student answers.

To achieve the first objective Retrieval Augmented Generation (RAG) should be explored to efficiently generate questions from the provided documents. This ensures that the generated questions are relevant for the school's courses.

The second objective involves evaluating the accuracy of grading student answers. This will be accomplished by creating a combined dataset from various grading datasets and conducting an evaluation of multiple Automatic Short Answer Grading (ASAG) systems. Most existing ASAG research does not evaluate systems outside of their specific dataset and domain, so this project will address that gap by providing a broader evaluation framework.

By developing this pipeline, the project aims to support students in the learning process through frequent self-testing, which research indicates helps memory retention greatly. Additionally, it ensures an accurate assessment of student progress, thereby enhancing the overall educational experience.

## B.2. Use of Generative AI Tools

This thesis utilised generative AI tools for text suggestions and reformulation, with both ChatGPT[12] and Grammarly[13] employed to suggest text formulations, improve clarity and coherence, check for grammatical errors, refine language, and ensure overall readability.

---

[12]Available online: **https://chatgpt.com**
[13]Available online: **https://grammarly.com**